\documentclass{article}

\PassOptionsToPackage{numbers, compress}{natbib}

\usepackage[preprint]{neurips_2023}

\usepackage[T1]{fontenc}    
\usepackage{hyperref}       
\usepackage{url}            
\usepackage{booktabs}       
\usepackage{amsfonts}       
\usepackage{nicefrac}       
\usepackage{microtype}      
\usepackage{xcolor}         
\usepackage{soul}
\usepackage{graphicx}

\newcommand{\re}[1]{{\color{black}#1}} 
\newcommand{\zs}[1]{{\color{black}#1}} 
\newcommand{\gh}[1]{{\color{black}#1}} 
\newcommand{\ys}[1]{{\color{black}#1}} 

\title{Reboost Large Language Model-based Text-to-SQL, Text-to-Python, and Text-to-Function \\
- with Real Applications in Traffic Domain}

\author{%
  Guanghu Sui$^\dagger$\\
  Sensetime Research\\
  Shanghai, China \\
  \texttt{suiguanghu@sensetime.com} \\
  \And
  Zhishuai Li$^\dagger$ \\
  SenseTime Research\\
  Shanghai, China \\
  \texttt{lizhishuai@sensetime.com}\\
  \And
  Ziyue Li\\
  University of Cologne\\
  Cologne, Germany\\
  \And
  Sun Yang\\
  Peking University\\
  Beijing, China\\
  \And
  Jingqing Ruan\\
  Chinese Academy of Sciences\\
  Beijing, China \\
  \And
  Hangyu Mao\\
  Sensetime Research\\
  Shanghai, China \\
  \And
  Rui Zhao\\
  Qing Yuan Research Institute\\
  Sensetime Research\\
  Shanghai, China \\
  \texttt{zhaorui@sensetime.com}\\
}

\begin{document}

\maketitle
\def\thefootnote{$\dagger$}\footnotetext{These authors contribute equally to this work.}\def\thefootnote{\arabic{footnote}}

\begin{abstract}
  \gh{Previous state-of-the-art (SOTA) method achieved a remarkable execution accuracy on the Spider dataset, which is one of the largest and most diverse datasets in the Text-to-SQL domain. However, during our reproduction of the business dataset, we observed a significant drop in performance. We examined the differences in dataset complexity, as well as the clarity of questions' intentions, and assessed how those differences could impact the performance of prompting methods.  Subsequently, We develop a more adaptable and more general prompting method, involving mainly query rewriting and SQL boosting, which respectively transform vague information into exact and precise information and enhance the SQL itself by incorporating execution feedback and the query results from the database content. In order to prevent information gaps, we include the comments, value types, and value samples for columns as part of the database description in the prompt. Our experiments with Large Language Models (LLMs) illustrate the significant performance improvement on the business dataset and prove the substantial potential of our method. In terms of execution accuracy on the business dataset, the SOTA method scored 21.05, while our approach scored 65.79. As a result, our approach achieved a notable performance improvement even when using a less capable pre-trained language model. Last but not the least, we also explore the Text-to-Python and Text-to-Function options, and we deeply analyze the pros and cons among them, offering valuable insights to the community.}

\end{abstract}

\section{Introduction}
\zs{Nowadays, growing volumes of data are structured and stored in databases, where the structured query language (SQL) serves as a crucial facility to communicate. Nevertheless, for those who are not well-versed in SQL, composing code to manage and operate databases is not effortless.
Converting natural language (NL) text to SQL queries \cite{guan2019improved}, Text-to-SQL (Text-to-SQL), as shown in Fig. \ref{fig:intro}, provides interfaces that enable users to interact with databases in a natural language manner, which contributes to facilitating the useability of database systems and improving the efficiency of data analysis\cite{deng2022recent}.
Fundamentally, it falls within the domain of machine question-answering, related to the application of semantic parsing\cite{nguyen2020pilot} techniques. Therefore, over the years, Text-to-SQL has evolved into a prominent research focus within the natural language processing (NLP) community\cite{wang2019rat,wang2020text,song2022speech}.

\begin{figure}
    \centering
    \includegraphics[width=0.9\columnwidth]{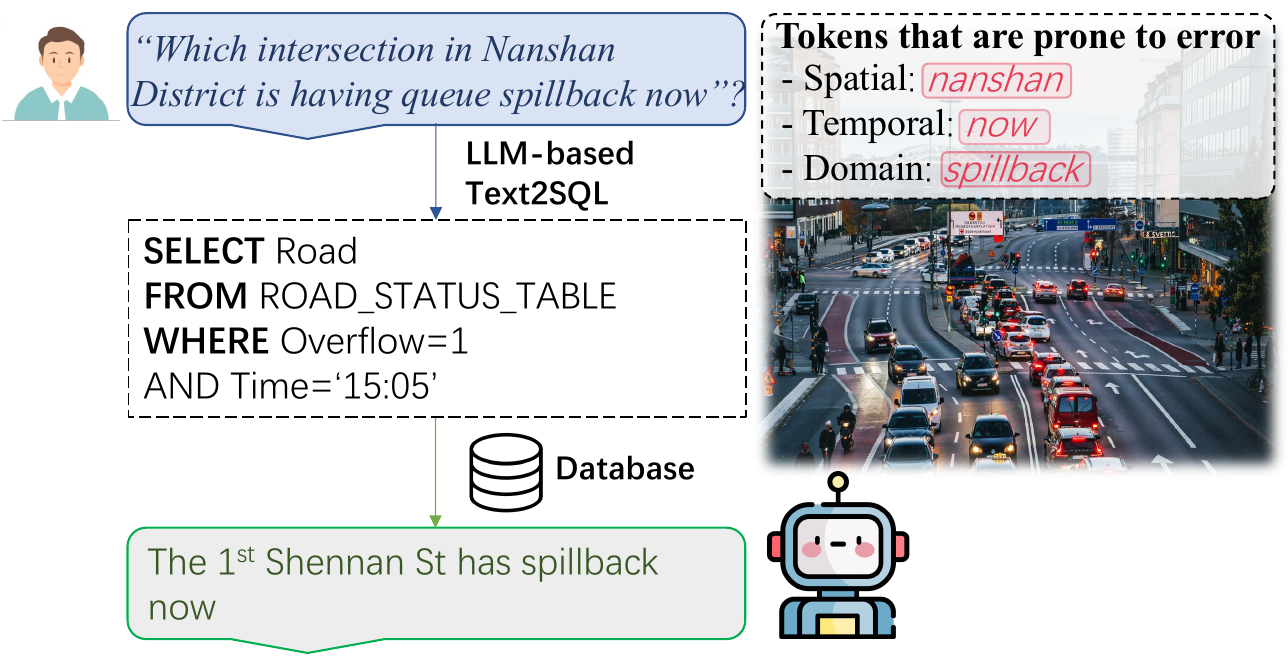}
    \caption{LLM-based Text-to-SQL model translates NL to database query and gives the proper answer}
    \label{fig:intro}
\end{figure}

The generalized avenues of Text-to-SQL can be summarized as understanding-then-generation.
Recent advances in Text-to-SQL are \re{further empowered (1) by \textbf{customizing deep language models} such as BERT \cite{devlin2018bert}, and (2) as the most recent trend, by \textbf{leveraging large language models (LLMs)}' strong understanding and generating power, such as GPT-4 \cite{openai2023gpt4} and PaLM-E \cite{driess2023palm}}. 
The former dominated the state of the arts for a long time, among which the sequence-to-sequence scheme demonstrated salient performance, such as \re{BRIDGE\cite{lin2020bridging} (BERT-based), RASAT\cite{qi2022rasat} (relation-embedding with Transformer encoder), RESDSQL\cite{li2023resdsql} (BART\cite{lewis2019bart}-based, with a schema-linking-ranking-enhanced encoding and skeleton-parsing-aware decoding), Graphix-t5\cite{li2023graphix} (T5\cite{xue2020mt5}-based, with specialized graph-aware layers.)}. 
However, the models may struggle to understand the meaning of terminology in an unseen domain, and the polysemy-related queries from different contexts complicate and confuse the code generation phase.
The latter, LLM-based Text-to-SQL methods, call the off-the-shelf LLMs to infer the results by prompting NL query and database information. 
Benefiting from the impressive natural language understanding (NLU) capabilities, LLMs-based models can easily understand the intents of the querier. Coupled with the dedicated design of the LLMs interaction, the latter has become the most advanced solution. Representatively, DIN-SQL\cite{pourreza2023din} broke down the task into sub-problems, each of which was solved by interacting with GPT-4, to significantly improve the performance. TPTU proposed to use a single agent and a set of sequential agents to solve simple SQL and complex nested SQL generation, respectively, based on various LLMs as backbone \cite{ruan2023tptu}.
}

\zs{
However, the dilemma of LLMs-based models is still evident: 
(1) Insensitivity to domain-specific words, generating incomplete conditions or codes (2) Constraints on limited contextual tokens, making it challenging to encode sufficient contextual information into LLMs, particularly when dealing with extensive databases (3) Buggy code generated from LLMs, resulting in SQL execution failures or incorrect results. As shown in Fig. \ref{fig:intro}, in the specific domain of traffic management, information such as spatial-temporal and domain knowledge can be ignored.
}

\zs{In response to the challenges, we elaborate a novel LLMs-based framework that decomposes the Text-to-SQL task into four components: (a) query rewriting, which targets the user's implicit intention; (b) explain-squeeze schema linking, which specifies the tables and columns in the database that correspond to the phrase in the given query (c) SQL generation, which aims to initial the query SQL, and (d) SQL boosting, which improves SQL code with the execution by the bug tracebacks to prompt LLMs repeatedly. Each one communicates with LLMs and gets their feedback as a result separately. 
Different from existing methods, it is designed for more practical and real business scenarios, with the following attributes: (1) users' questions are more implicit and domain-specific, (2) the database is composed of tables with more columns, and (3) most crucially, a lot of in-depth or even niche domain knowledge involved.
}
\gh{
For example, the Spider dataset \cite{yu2018spider} averages 5.4 columns per table, whereas our business dataset, \textit{CTraffic}, boasts an average of 26.2 columns per table, approximately five times Spider's.

The main contributions of our work are threefold:
\begin{itemize}
    \item We propose an elaborated Text-to-SQL framework that decomposes the task into four components, including query writing, explain-squeeze schema linking, SQL generation, and SQL boosting, each of which can interact with LLMs separately and alleviate their dilemma when encountering practical and real business scenarios. 
    \item To give a glimpse of the table structure and values, we include the comments, value types, and value samples of columns as auxiliary descriptions of the database in the prompt.
    \item Several experiments are conducted to validate the framework's superiority. The experimental result illustrates the effectiveness of our proposed method.
\end{itemize}}

\section{Related Work}
Two streams of related works are reviewed, including the customized learning-based and LLM-based methods.

\subsection{Customized learning-based models}
\zs{Earlier, Text-to-SQL was conducted rule-based or template-based approaches\cite{li2014constructing,mahmud2015rule}, which are well-suited for simple queries and size-limit databases. With the advancement of deep learning, especially the emergence of NLP for coding\cite{svyatkovskiy2020intellicode}, deep models have come into the spotlight. The customized deep models are more powerful, benefiting from their premising ability on representation and prediction, among which sequence-to-sequence scheme demonstrates salient performance. 
By encoding given queries in NL and the database schema, a decoder may follow to generate the target SQL.}

\zs{However, a tailor-trained model may sacrifice the generalization. Firstly, the limited corpus involved in the training stage leads to the insufficient ability for NL encoding. For an unexposed domain, the models may struggle to understand the meaning in terminology. Secondly, the polysemy-related information from different contexts makes the queries for different purposes similar, which complicates and confuses the code generation.
}
\subsection{Stimulating general LLM with prompting}
\zs{LLM have made significant gains in generalization capabilities, as they derive not only SQL code generation capabilities during pre-training, but also emerge with impressive language comprehension capabilities.}
\ys{\cite{liu2023comprehensive} carried out experiments on 12 benchmark datasets with varying languages, settings, and scenarios, revealing ChatGPT's robust capabilities in the field of Text-to-SQL.\cite{chang2023prompt} and \cite{nan2023enhancing} extend the concept of in-context learning to Text-to-SQL and underscore the effectiveness of their prompt design strategies, which enhance LLMs' performance. \cite{gao2023text} introduce a benchmark for Text-to-SQL empowered by Large Language Models (LLMs), and they evaluate various prompt engineering methods. Their work underlines the potential of open-source LLMs and the importance of token efficiency in prompt engineering. In the same vein, \cite{pourreza2023din} study the problem of decomposing a complex Text-to-SQL task into smaller sub-tasks, demonstrating how such a decomposition significantly improves the performance of Large Language Models (LLMs) in the reasoning process. They emphasize the potential of in-context learning and show that their approach consistently improves LLMs' few-shot performance, effectively pushing LLMs towards or beyond the state-of-the-art (SOTA).}

\zs{ General LLMs have demonstrated superior performance in SQL generation, but there is still room for improvement during the interaction, e.g., reasonable representation and appropriate feedback. Since there is no training involved, how to design a targeted pipeline to better stimulate NL understanding and SQL generation is a pivotal chapter.
}
\section{Methodology}
\begin{figure}[h]
\centerline{\includegraphics[width=0.99\textwidth]{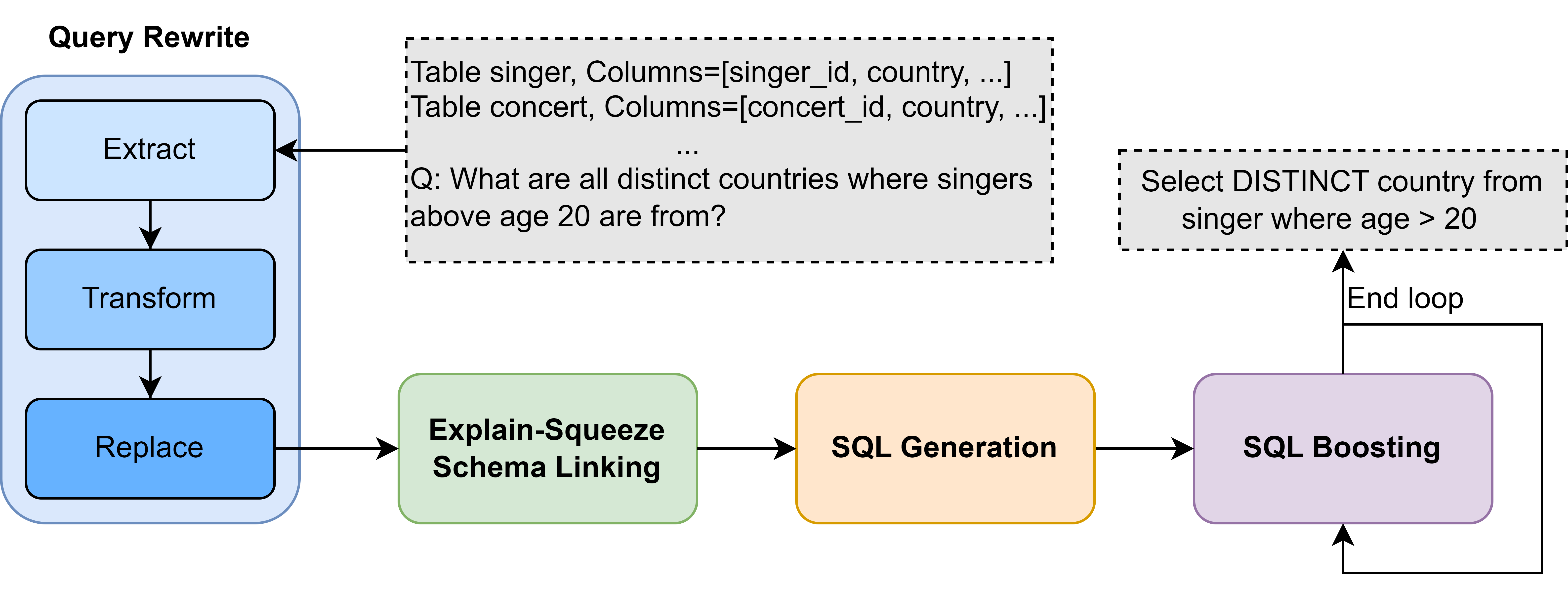}}
\caption{The overview of the proposed method: ReBoostSQL}
\label{fig:overview}
\end{figure}

\zs{In this section, we will elaborate on our dedicated approach to SQL generation. The overview is shown in Fig. \ref{fig:overview}, which is a cascaded framework with four modules: (a) query rewriting, (b) explain-squeeze schema linking, (c) SQL generation, and (d) SQL boosting. Each component interacts with the LLMs separately. Next, we detail their distinct inspirations and roles in the framework.}

\begin{figure}[t]
\centerline{\includegraphics[width=0.75\columnwidth]{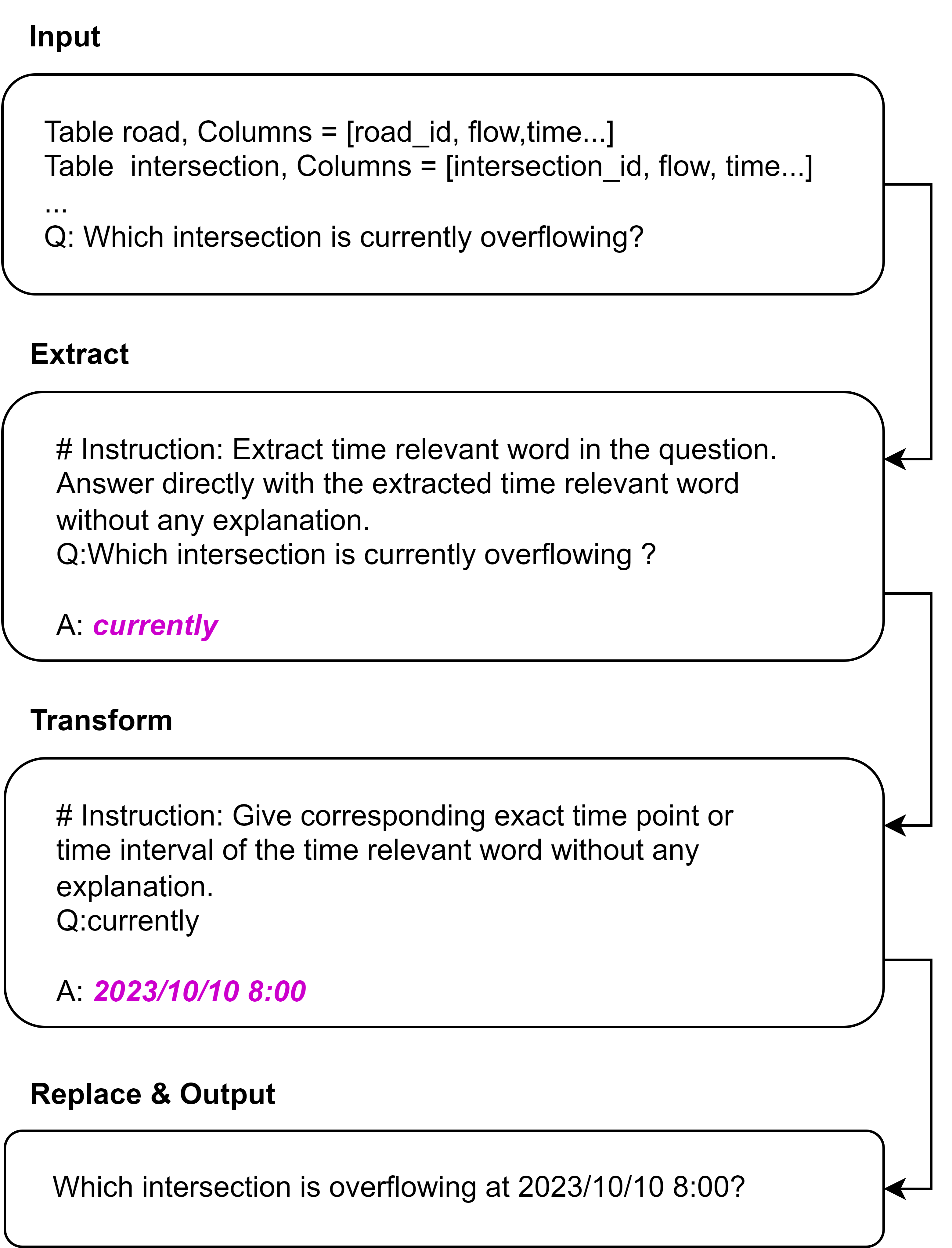}}
\caption{An illustration of Query Rewriting process with prompts of extract, transform and replace}
\label{fig:rewrite}
\end{figure}

\subsection{Query Rewriting}

\zs{Although LLMs demonstrate strong NLU capability, they are still insufficiently sensitive to particular information,  \re{including spatiotemporal background such as the current timestamp, location/area/district, and also some database- or domain-specific definitions}). For instance, when a question contains expressions such as `\textit{now}' and `\textit{just now}', LLMs usually fail to link the current system timestamp (assuming 2023-10-10 08:00) into the SQL condition and output queries like ``\textit{select ... where time=`2023-10-10 08:00'; }". This is unfavorable for the exact database retrieving.} \re{We observed the same phenomenon where LLMs can not understand our data's spatial and domain-specific terms.}

\gh{To solve the problem, we first rewrite the query, which intends to replace those vague words with more exact information, for example, a detailed time or a clear time interval instead of `\textit{now}', `\textit{just now}', etc. \re{This significantly complete the semantics of the original query.}

\re{As shown in Fig. \ref{fig:overview}}, the query rewriting mainly includes three steps: extracting, transforming, and replacing. (1) Extracting is the process of finding out the relevant vague words that need to be rewritten to get a clear representation of information by LLM. (2) Transforming is to convert the extracted word to \re{explicitly-defined} terms using the context or internal knowledge of the pre-trained language model. (3) Replacing forms the output with the vague words replaced by the explicitly defined terms.} \re{Fig. \ref{fig:rewrite} gives a demonstration of the rewriting step (rewriting temporal information).}

\begin{figure}[t]
    \centering
    \includegraphics[width=0.85\columnwidth]{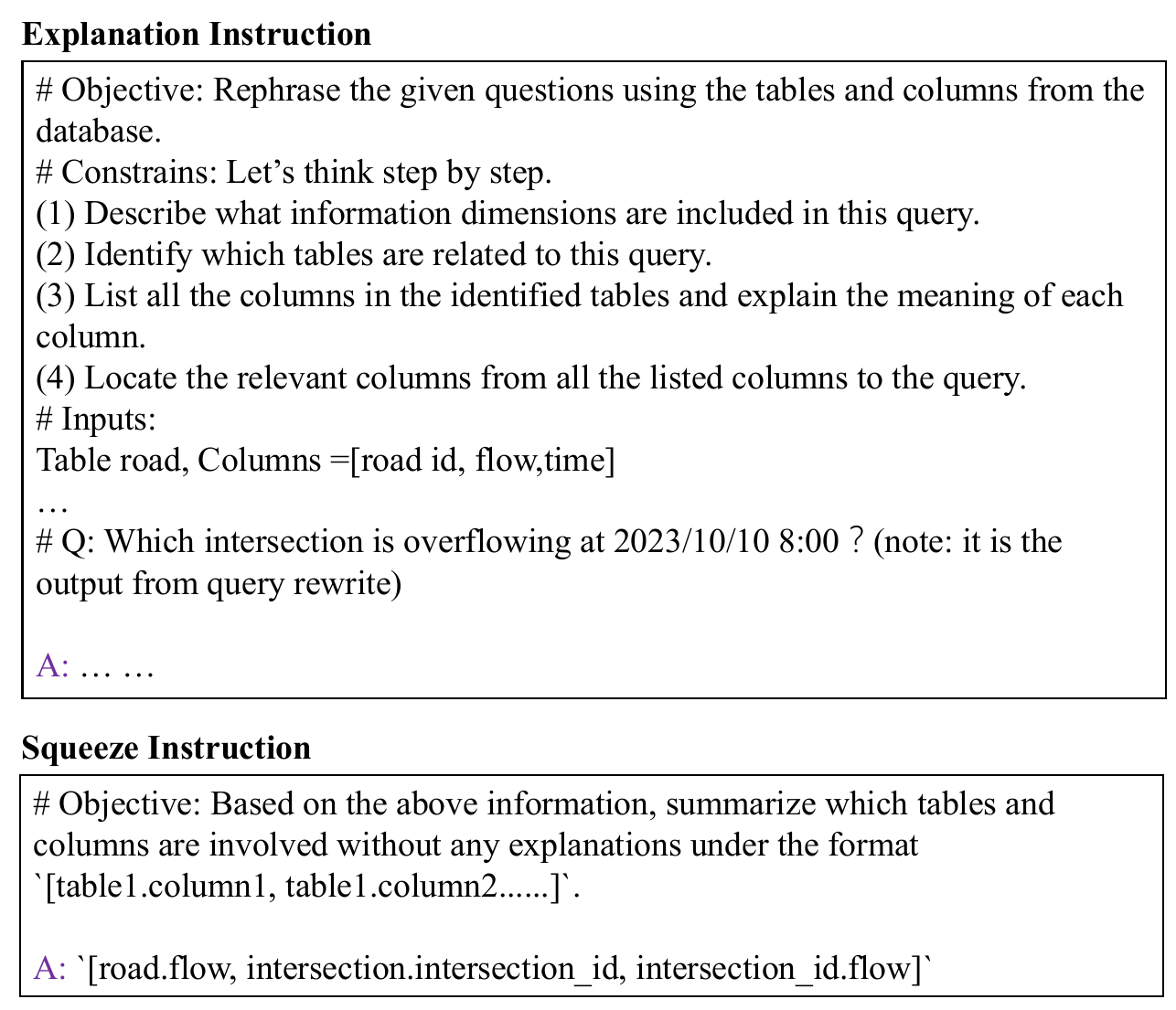}
    \caption{The illustration of the explain-squeeze schema linking pipeline.}
    \label{fig:sche_pip}
\end{figure}

\subsection{{Explain-Squeeze} Schema Linking}
\zs{For a large database, it is impractical to prompt all the table descriptions into the LLM and generate a response to the query directly due to the limited tokens. 
Therefore, schema linking is a plugin that serves as a preprocessing step before SQL generation.
Its objective is to specify the tables and columns in the database that correspond to the phrase in the given query. As in shown Fig. \ref{fig:sche_pip}, we devise an \textbf{explain-squeeze pipeline} for schema linking, where coarse database information inputs to the LLM and detailed descriptions exclusively for the selected tables and columns are generated by prompting it.

(1) Specifically, \textbf{in the explain phase}, we first find all the table-column pairs in the database, which are represented under ``\texttt{Table $1$, Columns=[column $1$, column $2$, ..., column $N$]; ...; Table $M$, ...;}" format. Coupled with them, we instruct the LLM to rephrase the query using tables and columns from the database step by step: 
\re{
\begin{itemize}
    \item Describe what information dimensions are included in this query.
    \item Identify which tables are related to this query. 
    \item List all the columns in the identified tables and explain the meaning of each column.
    \item Locate the relevant columns from all the listed columns to the query.
\end{itemize}
}
(2) Then, \textbf{in the squeeze phase}, we feed the response from the explanation to the LLM and request it to deliver the structured content of used tables and columns as the summarization for schema linking, for example ``\texttt{[Table1.column3, Table2.column1,...]}". } As a result, our explain-squeeze design significantly saves the tokens for the prompt and maintains the accuracy of schema linking.

\begin{figure}[t]
\centerline{\includegraphics[width=0.70\columnwidth]{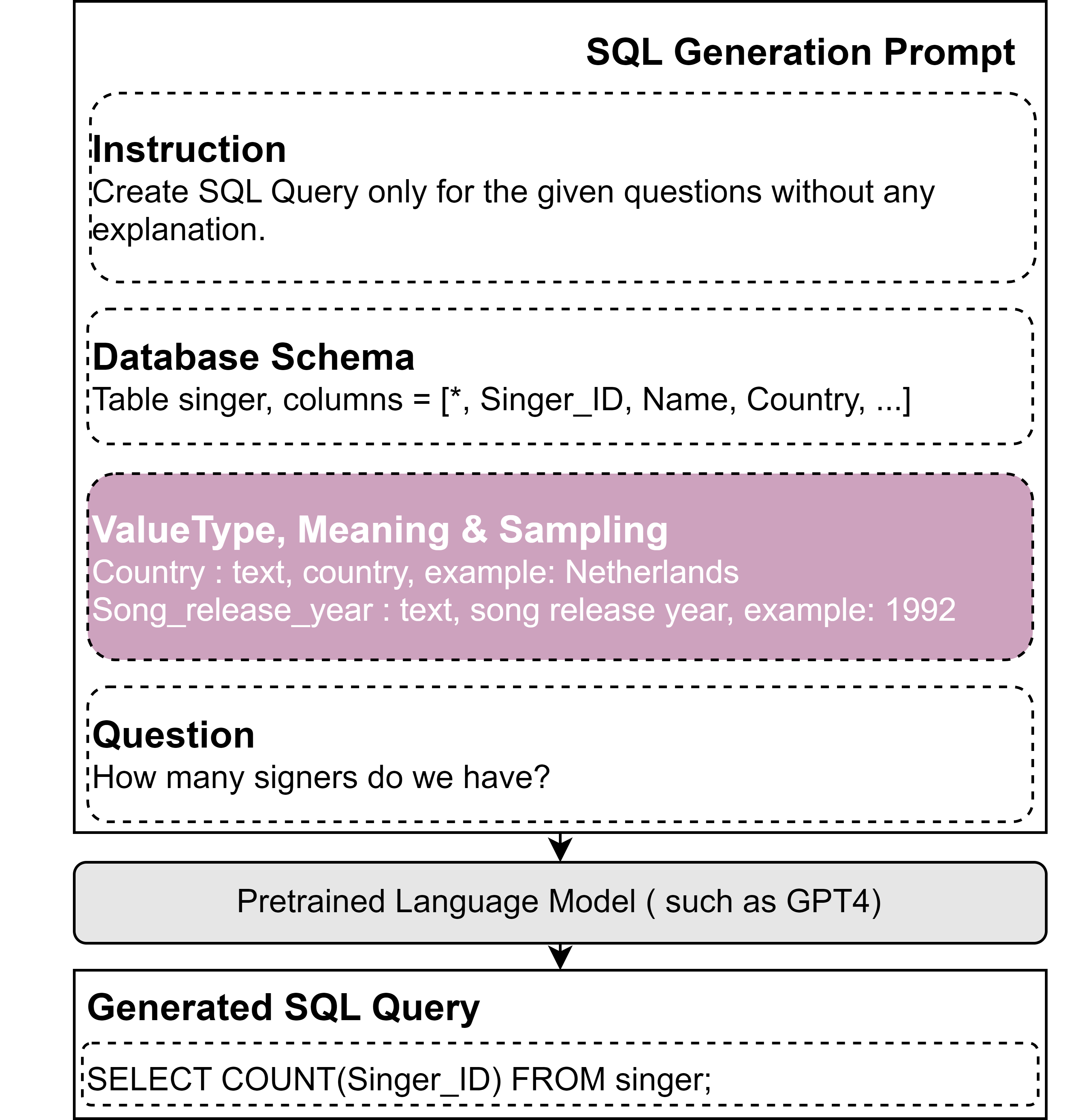}}
\caption{An example showing the process of SQL generation with `ValueType, Meaning and Sampling' highlighted.}
\label{fig:sqlgen}
\end{figure}

\subsection{SQL Generation}

\gh{SQL generation targets to obtain an initialized query SQL for SQL boosting module. Different from schema linking, which is to select tables and columns in limited dimensions, generating a sequence of words is nearly of unlimited dimension. More specifically, both the length of the sequence and each word of the sequence are also not fixed, and the word candidate is of large dimension. Theoretically, without a well Pre-trained LLM, the dimension of word candidates could be the whole dictionary length. All those lead to a high complexity for generating a well-defined SQL query. Even for humans, it is extremely hard to write once and obtain a correct SQL. Therefore, our goal is to generate a ready SQL query as an initialization of the final SQL query and as approximate as possible the ground truth. \re{Specifically, we used ChatGPT, GPT-4, and an anonymous LLM from a well-known company in the AI industry (confidentiality protected) as the generation model. More details can be found in Section \ref{sec:setting}.}

\re{It is worth mentioning that providing necessary and sufficient information is essential for LLM to ensure decent performance}. Otherwise, the lack of information may lead to the wrong choice of tables or columns, the wrong representation of the condition presented in the question, and the wrong value \cite{yang2023harnessing}. Usually, the prompt to generate SQL comprises three parts: instruction, database schema, and question, which is also generally the classification of the necessary and sufficient information mentioned above. For database schema, DINSQL \cite{pourreza2023din} simply uses the table name and column name as representation. However, during the reimplementation, we found that the table and column names are insufficient to describe the database and consequently generate robust SQL queries. There exist different types of mistakes, including wrong selection of tables or columns, fault representation of condition, and usage of value. For example, the date could be expressed in the format of `2023/10/10', or `2023-10-10', or even `10 October 2023'. Without information on the format of value, the LLM may select one randomly, and thus, the query result couldn't be ensured.

To address this issue, we use more information to describe the columns in the database, including value type, column meaning, and a sample. Fig. \ref{fig:sqlgen}  shows the components of the prompt for generating SQL, and the highlighted part is the additional information to describe the database.}

\begin{figure}[t]
\centerline{\includegraphics[width=0.99\columnwidth]{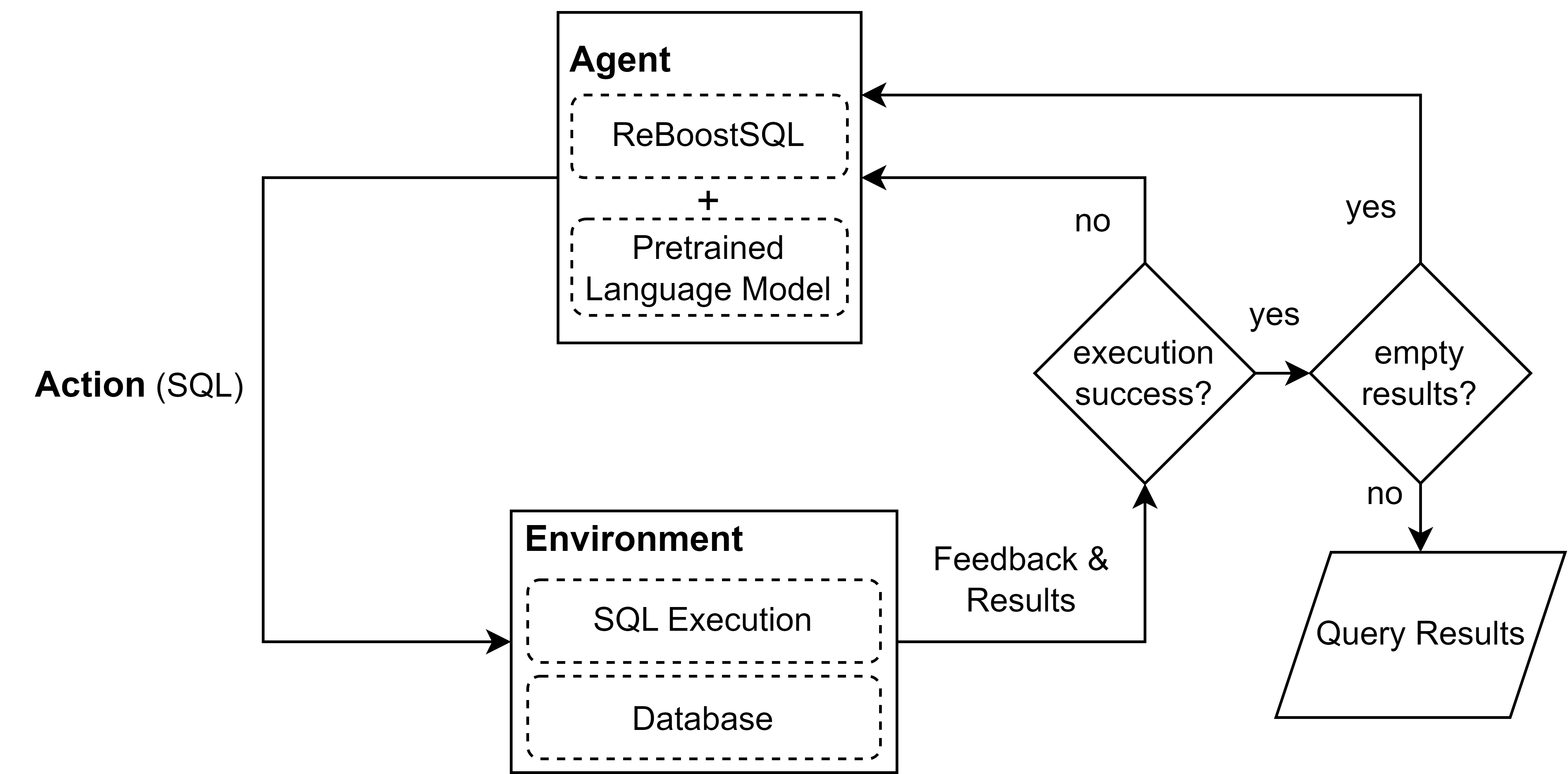}}
\caption{The process of SQL Boosting.}
\label{fig:sqlboost}
\end{figure}

\subsection{SQL Boosting}
\gh{Since human usually uses feedback to adjust the SQL query, it is natural to use that feedback for LLM to improve the SQL query by itself. 
Thus, we design the process of SQL Boosting, which improves the query SQL by execution feedback and query results as shown in \ref{fig:sqlboost}. Here, we use concepts of reinforcement learning to describe the process (without the implementation of reinforcement learning methods behind). The agent, composed of prompt method and LLM, is responsible for generating SQL. The generated SQL, which is action, is then fed into the environment. 
The prompt method is ReBoostSQL, in our case. Then, the environment, which is based on the database and an SQL compiler, would execute the SQL. The output of the environment contains two parts: feedback and query results. The feedback includes the label of execution status, which is `success' or `fail', and the error message if it encounters bugs. The query results are the required data in the database, which are the direct answer to the user intent question. \re{An important point is that the model works similarly in a way as reinforcement learning}, which uses observation and reward to update the agent. 

It is worth comparing SQL boosting with 2 existing methods: self-debugging \cite{chen2023teaching}  and self-correction \cite{pourreza2023din}. Self-debugging emphasizes the code explanation and uses the correctness statement inside the generated code explanation to judge whether to end the optimization loop. Our method simply relies on the feedback generated by the SQL executor and generally uses the condition that the query results are not empty as a termination condition. 
Meanwhile, self-correction depends on the pre-defined rules and has no interaction with SQL interpreters. Our approach has no pre-defined rules and heavily depends on the code interpreter environment.

The detailed process is shown in Fig. \ref{fig:sqlboost}. Precisely, we first simply use the success of execution as the ending condition of the boosting loop. Then, we found that this configuration would lead to empty query results since the success of execution could ensure syntax correctness but can't ensure semantic correctness. To make the results closer to ground truth, we set the condition that the query result is not empty as the ending condition and a max try of 3 times in case the empty result is truly positive.}

\section{Experiment}

\subsection{Dataset}

\begin{figure}[ht]
\centerline{\includegraphics[width=0.82\columnwidth]{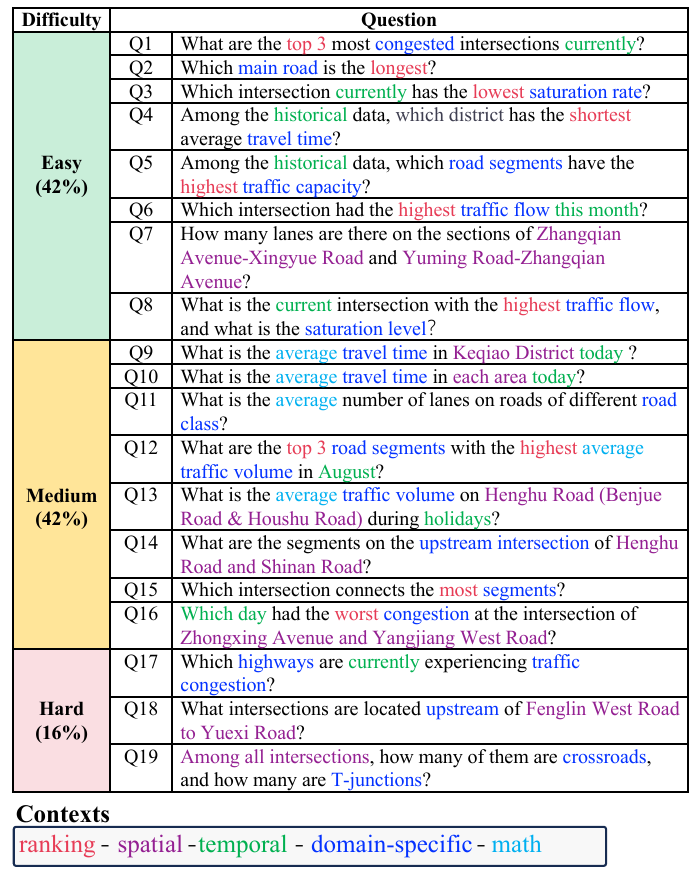}}
\caption{Selected 19 Questions related to \textit{CTraffic}, ranked into three classes based on difficulty: easy, medium, and hard \cite{pourreza2023din}}
\label{fig:question}
\end{figure}

\gh{As our goal is to solve practical business problems, we conduct the experiments directly on two business databases, with each database containing multiple tables and each table containing multiple columns. The datasets are across two domains, including traffic congestion and company management. Those databases are all provided by related companies and obtained from their existing real business scenarios, having run and worked for a long time. The questions are provided by business experts, and corresponding query SQLs are figured out and verified by professional SQL engineers.

Specifically, Fig. \ref{fig:question} demonstrates the questions in \textit{CTraffic}. We have manually categorized them into three groups based on the query complexity, the number of tables involved, and the presence of aggregation operations: (1) Easy: query on a single table without aggregation function; (2) Medium: query on a single table with aggregation functions (e.g., sum, avg); and (3) Hard: cross-table query, which may involve multiple tables and may require aggregate functions.

The distributions of the two databases are described in Table~\ref{tab1}: the \textit{CTraffic} consists of 19 questions, while the \textit{CompanyZ} consists of 17 questions. The \textit{CTraffic} is the abbreviation of Customized Traffic. The \textit{CompanyZ} refers to the operational data of an anonymous company Z. For \textit{CTraffic}, there are 5 tables and 26.2 columns per table on average. For \textit{CompanyZ}, there are 5 tables and 17.8 columns per table on average. 
In contrast, Spider, one of the biggest datasets in the field of Text-to-SQL, has only 5.4 columns in each table on average, which is far less than the two practical datasets \textit{CTraffic} and \textit{CompanyZ}. 
In fact, with more columns in each table, it is getting harder to figure out the correct columns to select and the complex relationship between those columns. However, in reality, especially in the domain of information technology, with the increase in query demands and information, the table is intended to have more columns in order to decrease the conversion of tables. Those two business databases are proofs. Regarding the average number of tables in each database, \textit{CTraffic}, \textit{CompanyZ}, and Spider remain to be the same level, which is approximately five.\\
}

\begin{table}[h]
\caption{Details of the databases}
\begin{center}
\begin{tabular}{p{1.6cm}p{1.6cm}p{1.6cm}p{1.6cm}p{2.5cm}}
\hline 
Dataset&\#Questions&\#Databases&\#Tables/DB&\#Columns/Table \\
\hline
\textit{CTraffic} & 19 & 1  &5 &26.2\\
\textit{CompanyZ} & 17 & 1 & 5  & 17.8\\
\hline
Spider & 10,181 & 200 & 5.1 & 5.4 \\
\hline
\\
\end{tabular}
\end{center}
\label{tab1}
\end{table}

\subsection{Metrics}
\gh{To validate the performance of our method, execution accuracy (EX) is calculated as the evaluated criteria, which
involves comparing the execution results of the predicted SQL query with those of the ground truth SQL query. 
This provides a more precise evaluation of the model's performance, with a particular emphasis on practical applications.

}

\subsection{Settings}
\label{sec:setting}
\gh{We evaluated the proposed framework by two LLMs: ChatGPT and GPT-4, which are the largest and most advanced off-the-shelf LLMs. 
To illustrate the relative performance of our method, 
we select DINSQL, one of the SOTA approaches, as a baseline.
It is worth mentioning that DINSQL adopts the few-shot setting, while our method adopts a zero-shot setting. We believe that the internal knowledge is enough for LLMs to figure out the Text-to-SQL tasks; thus, we decided to use a zero-shot setting.

To evaluate the generalized ability of ReBoostSQL, we also designed a comparison experiment to see if it could bring benefits to more  LLMs, not only the most advanced LLMs like ChatGPT or GPT-4. Therefore, we adopt an available released LLM. For the sake of confidentiality, we use `Anonymous LLM' to name it. The experiment was conducted twice, whether using ReBoostSQL or not.

In order to get a stable result, we set low temperatures for all LLMs. More specifically, for GPT-4 or ChatGPT \footnote{\url{https://platform.openai.com/docs/guides/gpt}}, the temperature is 0, while for `Anonymous LLM', the temperature is set to be 0.1.}

\subsection{Results}
\begin{figure}[t]
    \centering
    \includegraphics[width=0.8\columnwidth]{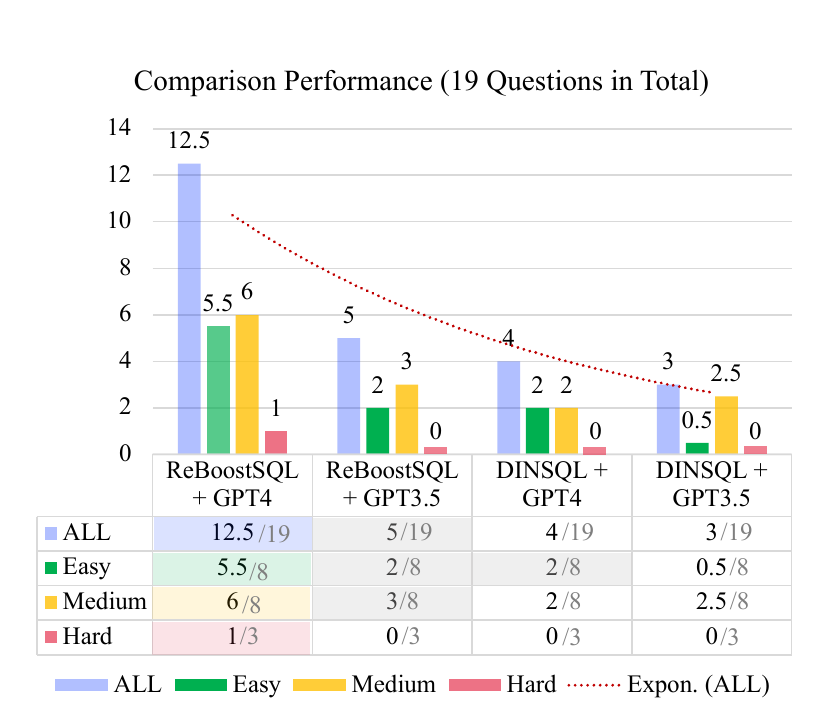}
    \caption{\re{Comparison on the performance of benchmarks over different levels of difficulties based on \textit{CTraffic} data, with the best highlight in colors, and the second best highlighted in grey}}
    \label{fig:enter-label}
\end{figure}

\gh{As shown in Table \ref{tab2}, our method sets a new ground, achieving the highest execution accuracy using GPT-4 and the second-highest execution accuracy using ChatGPT comparing the SOTA method DINSQL. The detailed execution results on each group are illustrated in Fig. \ref{fig:enter-label}. More specifically, the execution accuracy of ReBoostSQL with GPT-4 is nearly 3 times higher than the SOTA method, which is DINSQL with GPT-4. Interestingly, ReBoostSQL with ChatGPT achieves higher accuracy than DINSQL with GPT-4, illustrating that a good prompting method could compensate for the disadvantage of LLMs.}

\renewcommand{\arraystretch}{1.4}
\begin{table}[t]
\caption{Performance comparison on \textit{CTraffic}}
\begin{center}
\begin{tabular}{cc}
\hline 
Approaches & Execution Accuracy  \\
\hline
DINSQL \cite{pourreza2023din} + ChatGPT & 0.1579 \\
DINSQL \cite{pourreza2023din} + GPT4 & 0.2105 \\
ReBoostSQL (ours) + ChatGPT & 0.2632 \\
ReBoostSQL (ours) + GPT4 & \textbf{0.6579} \\
\hline
\\
\end{tabular}
\end{center}
\label{tab2}
\end{table}

\gh{To exclude the influence of LLM, we further analyzed the performance of our proposed method on queries with a less capable LM (a coding model, with only 5 billion parameters) on the other dataset, which is \textit{CompanyZ}. Table~\ref{tab3} presents the performance of our proposed method compared to a basic prompting on the \textit{CompanyZ} dataset. The accuracy of anonymous LLM with ReBoostSQL is nearly 2 times the accuracy of pure LLM, which demonstrates that our method worked well even for a less capable LLM.}

\begin{table}[t]
\caption{Performance comparison on \textit{CompanyZ}}
\begin{center}
\begin{tabular}{cc}
\hline 
Approaches & Execution Accuracy \\
\hline
Anonymous LLM              & 0.21 \\
Anonymous LLM + ReBoostSQL & \textbf{0.44} \\
\hline
\\
\end{tabular}
\end{center}
\label{tab3}
\end{table}

\section{Extended Results 1: How about Text-to-Python (Text2Python)?}
\subsection{Introduction of Text2Python}
Given that Python is a more universal and commonly used programming language than SQL, we can reasonably anticipate that the LLMs have been trained on high-quality Python codes. Therefore, it is possible that the LLMs may be more efficient in Python code generation than SQL. In response to this, we investigate the feasibility of using Python code for the database querying and question-answering task. This task is also termed Text-to-Python, which means generating Python code to interact with data according to users' questions.

\begin{figure}
    \centering
    \includegraphics[width=0.8\textwidth]{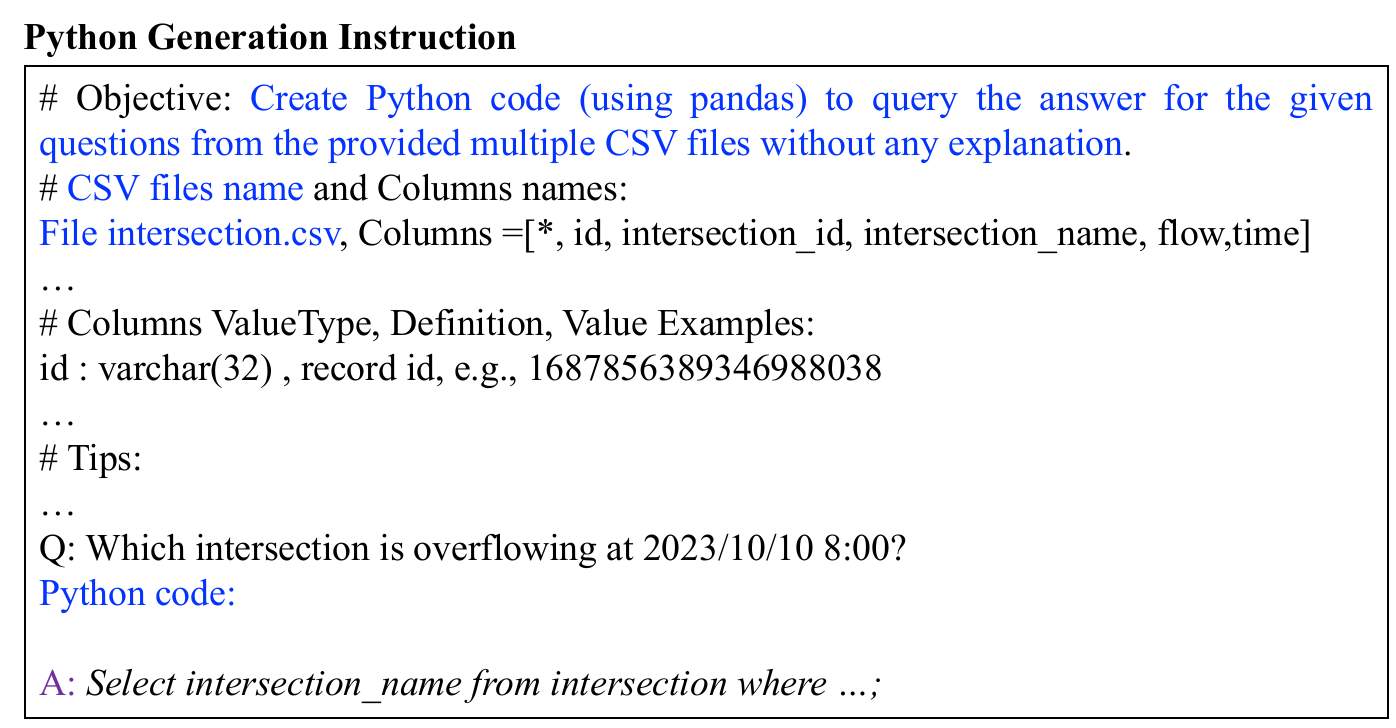}
    \caption{An example of the instruction for Text-to-Python in the code generation stage of the ReBoost, where the text in blue indicates the modifications compared to the SQL generation instructions.}
    \label{fig:text2py_inst}
\end{figure}

We extend the presented ReBoost framework to the Python (especially Pandas) code generation task, in which the proposed four modules are still applied. The dedicated modifications for the task lie in the SQL generation component: (1) The tables from the database as stored as separate files in comma-separated value (CSV) format and Pandas is used to load the selected file from the schema linking module.
(2) For the instructions that prompt LLMs, the ``Create SQL code" is replaced with ``Create Python code (using Pandas)". Fig. \ref{fig:text2py_inst} demonstrates the instructions used for the SQL generation module, in which the modifications are marked in blue.

\subsection{Preliminary Results}
The performance of Text-to-Python is examined again under the EX criteria. The results on the \textit{CTraffic} dataset are listed in Table \ref{tab:text2py}. 
It can be seen that, compared to Text-to-SQL, using Python to retrieve knowledge from the database achieves slightly inferior performance. 
Moreover, Text-to-Python consumes a greater number of generation tokens since the Python codes are usually longer than SQL text (even 4$\sim$5 times). Therefore, for the LLMs with size-limit tokens, Text-to-Python may not be the prioritized option to be adopted for database retrieving. 
\subsection{Analysis}
We also investigate the failed cases in the Text-to-Python processes. These cases typically involve two primary types:  (1) data type errors and (2) ``Hallucinations'' in LLMs. The former occurs when LLMs struggle to correctly discern the variable type of columns, e.g., \textit{str} are misinterpreted as \textit{int}, raising type conversion execution errors.
The latter is a well-known challenge in LLMs. Specifically, in the context of Text-to-Python, it may manifest as fabricated data rather than data extracted from the provided CSV files. 

\begin{figure}
    \centering
    \includegraphics[width = \columnwidth]{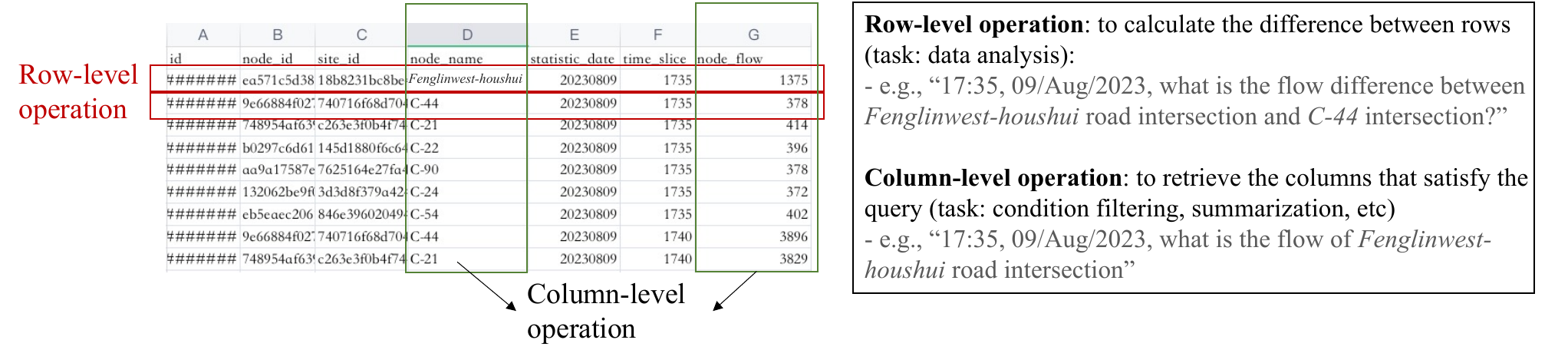}
    \caption{The comparison of row-level operation and column-level operation}
    \label{fig:row_vs_colomn}
\end{figure}

The differences between Text-to-SQL and Text-to-Python solutions can be concluded coupled with the oriented scenarios of the SQL and Python programming languages: 
Text-to-SQL is particularly well-suited for operations like conditional filtering and data grouping, i.e., extracting the column values from a table conditioned on the contexts (denoted as column-level operation in Fig. \ref{fig:row_vs_colomn}). 
It is suitable for applications with real-time updates in large-scale databases. 
On the other hand, Text-to-Python emphasizes data analysis features with more complex demands, such as basic math calculation and even visualization (denoted as row-level operation in Fig. \ref{fig:row_vs_colomn}), and is typically employed to work offline.  
To sum up, when it comes to database retrieval tasks, SQL can accomplish similar functions as Python and even outperforms it in some cases while imposing a lighter token load on LLMs.

\renewcommand{\arraystretch}{1.4}
\begin{table}[ht!]
\caption{Performance comparison on \textit{CTraffic} with Text-to-Python}
\begin{center}
\begin{tabular}{cc}
\toprule
Approaches & Execution Accuracy  \\
\midrule
Text-to-Python + ChatGPT & {0.368} \\
Text-to-SQL + ChatGPT & \textbf{0.421}\\
\bottomrule
\end{tabular}
\end{center}
\label{tab:text2py}
\end{table}

\section{Extended Results 2: How about Text-to-Function (Text2Function)?}


The application of LLMs, through designing prompt engineering, has demonstrated substantial promise and efficacy in the Text2SQL domain. However, this effectiveness seems to wane when confronted with the challenge of producing complex, nested, and elongated SQL statements, pushing the generative capacities of these models to their limits. One notable issue in leveraging LLMs is the emergence of ``Hallucinations'', which means generating completely fabricated information, such as incorrect table names, column names, etc.
In light of these challenges, we designed a novel methodology, Text2Function. Text2Function exhibits preliminary success and potential in handling the sophisticated aspects of Text2SQL generation tasks while mitigating the issues attributed to the ``Hallucinations'' in LLMs.

\subsection{Introduction of Text2Function}

Text2Function is designed to make SQL dataset queries more efficient and user-friendly. Its main goal is to create customized problem templates for SQL datasets. 
LLMs are responsible for organizing problem templates and giving specific parameter instances, making it easier to solve original text problems.
In this approach, function templates are closely linked to SQL query statements. This means that each question template matches a certain kind of SQL query, making the conversion from text to SQL more accurate and reliable. We have created four main categories of function templates to cover different types of SQL queries, helping to improve the system’s ability to handle various SQL dataset challenges effectively and accurately.

\begin{figure}[ht!]
    \centering
    \includegraphics[width=1.0\columnwidth]{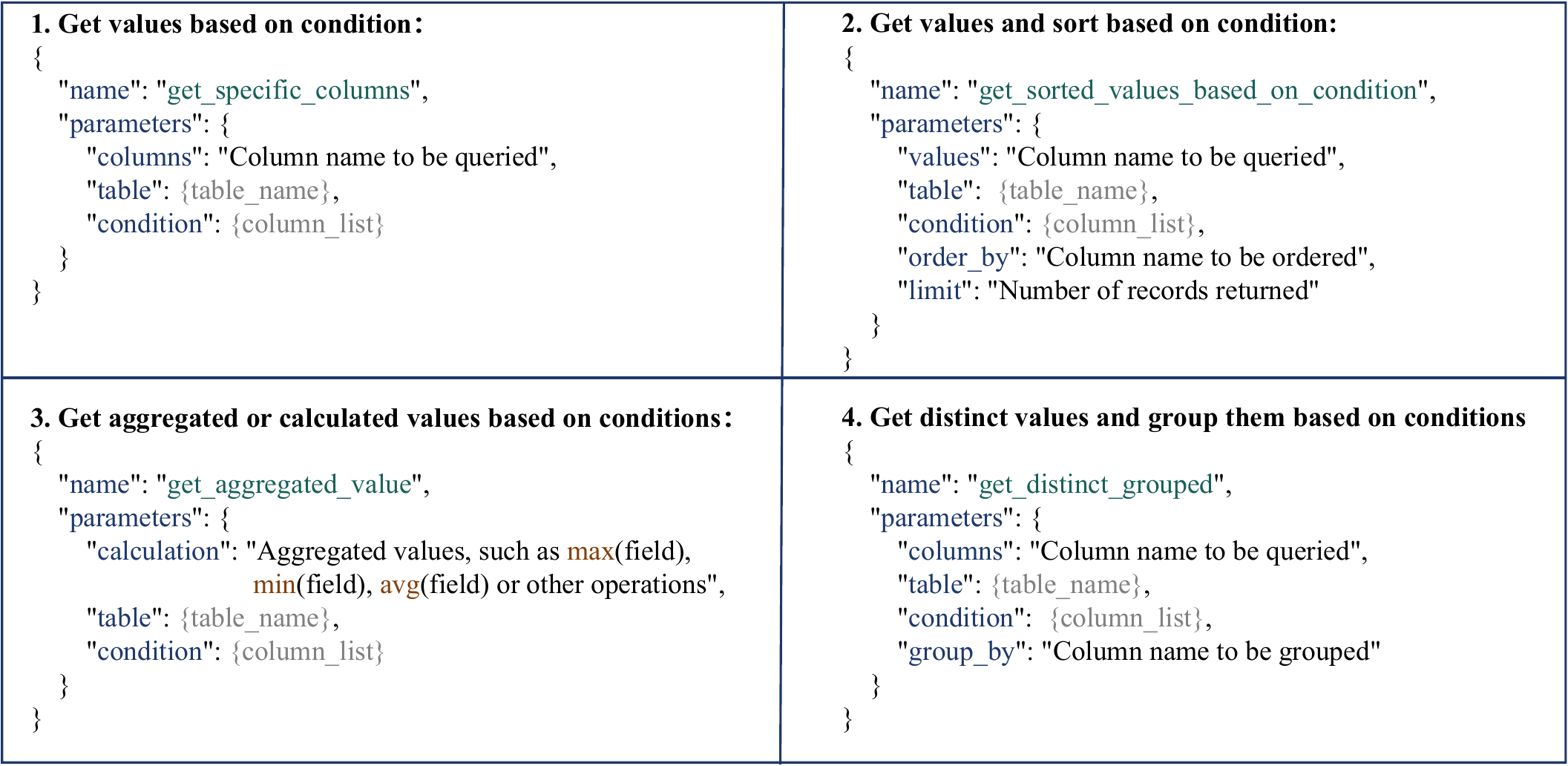}
    \caption{Four function templates for Text2Function.}
    \label{fig:text2func-templates}
\end{figure}

In Figure~\ref{fig:text2func-templates}, we present four principal categories of function templates, each designed to encapsulate various aspects and complexities of SQL queries. 
\begin{itemize}
    \item \textit{get\_specific\_columns(columns, table, condition)}: It is designed to execute queries that retrieve specific column values, utilizing conditions parsed from the user's query.
    \item \textit{get\_sorted\_values\_based\_on\_condition(values, table, condition, order\_by, limit)}: It is designed to retrieve specific column values, ordered according to certain conditions, and allows for setting a limit on the number of returned results. 
    \item \textit{get\_aggregated\_value(calculation, table, condition)}: It is designed to execute queries that retrieve aggregated or calculated values from specific columns, enabling a more analytical and synthesized view of the data. 
    \item \textit{get\_distinct\_grouped(columns, table, condition, group\_by)}: It is designed to execute queries that retrieve values and group them based on specified conditions or attributes. 
\end{itemize}

Thus, when given a user query, Text2Function will strategically schedule and employ LLMs, prompted with these specifically designed function templates. This process leads to the generation of final answers, utilizing appropriate functions to adeptly solve and address the query problems, ensuring that the user receives accurate and comprehensive solutions.

\subsection{Preliminary Results }

In this section, we initially assess the performance by conducting tests on \textit{CompanyZ}, using GPT-3.5 and GPT-4 as baselines. Following this, we enhance this baseline by incorporating our Text2Function module, enabling us to evaluate its potential and effectiveness in improving query results. The empirical outcomes of these assessments, reflecting the comparative performance and potential enhancements brought about by the Text2Function module, are systematically presented in Table~\ref{tab:text2function}.

\renewcommand{\arraystretch}{1.4}
\begin{table}[ht!]
\caption{Performance comparison on \textit{CompanyZ} with Text2Function}
\begin{center}
\begin{tabular}{cc}
\hline 
Approaches & Execution Accuracy  \\
\hline
ChatGPT & 0.4117 \\
GPT4 & 0.5294 \\
Text2Function + ChatGPT & \textbf{0.6470} \\
Text2Function + GPT4 & \textbf{0.7647} \\
\hline
\\
\end{tabular}
\end{center}
\label{tab:text2function}
\end{table}

\subsection{Analysis }

In our experiments, various approaches were assessed for execution accuracy. The results reveal that the "Text2Function + LLMs" combination yields higher accuracy, outperforming baselines. 

Our preliminary analysis suggests that compared to directly generating SQL, more granular, customized functions are advantageous for LLMs in reducing the ``Hallucinations'' phenomenon. Direct SQL generation poses a considerable burden on LLMs, with more complex, nested SQL statements requiring higher generative capacities. Customized functions can mitigate common errors and the ``Hallucinations'' phenomenon, alleviating the generative burden on LLMs. However, increased granularity in function customization leads to higher manual costs and difficulties in large-scale API selection for LLMs, and may cause overfitting, making the functions overly specialized and less generalizable to new tasks.

For solutions like Text2Function, coarser functions cover multiple dataset functions but don’t alleviate the LLMs’ generative burden. A finer granularity exacerbates the mentioned issues. Future work to find a balance in function granularity is crucial for unleashing the full potential of this approach.

\begin{figure}[h]
    \centering
    \includegraphics[width = \columnwidth]{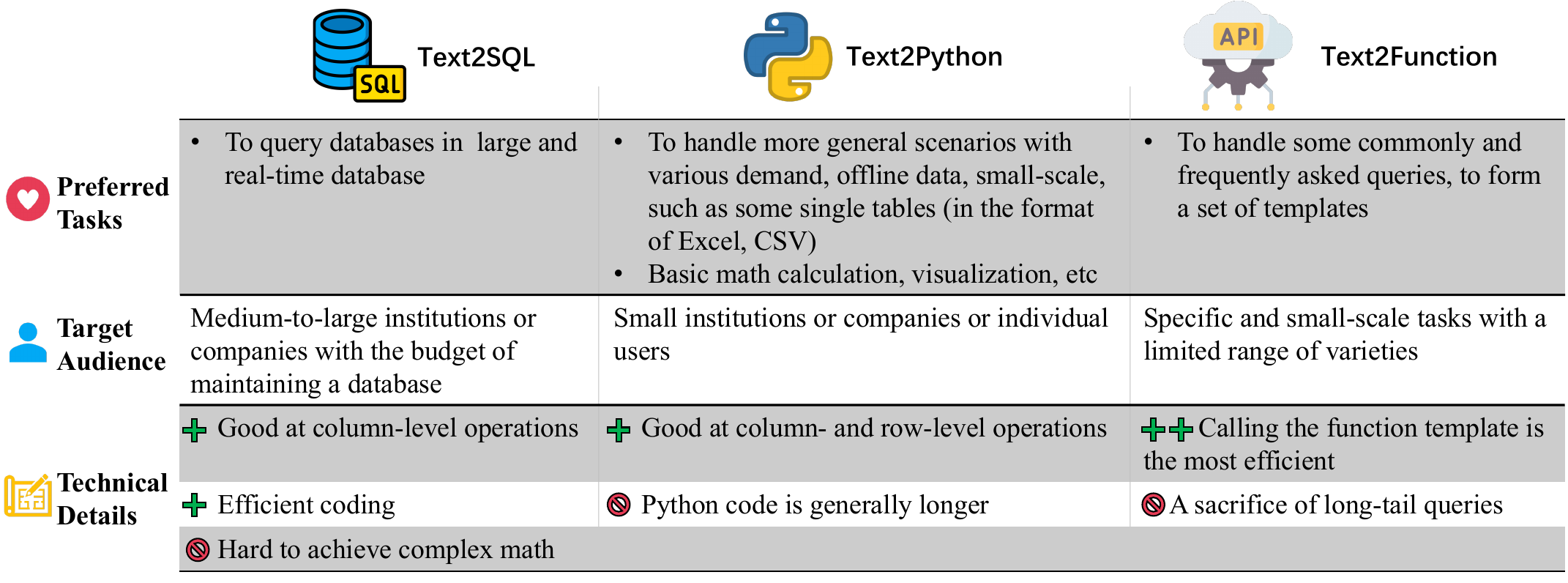}
    \caption{The comparison of Text-to-SQL, Text-to-Python, and Text-to-Function}
    \label{fig:compare}
\end{figure}

Overall, we summarize the comparison of Text-to-SQL, Text-to-Python, and Text-to-Function in Figure. \ref{fig:compare}.

\section{Conclusion}

\gh{The utilization of prompting has empowered large language models to exhibit remarkable performance across various NLP tasks in diverse domains, all without the necessity for an extensive training dataset. When applied to Text-to-SQL problems on large-scale datasets such as Spider, prompting has showcased the appreciated performance and profound potential. 
However, the challenge persists due to disparities in dataset distribution and human behaviors, requiring a higher level of the generalization ability of the prompting method in a real-world deployment.

In this study, we develop a novel method that mainly integrates query rewriting and SQL boosting to tackle these challenges. Compared to the SOTA method, our proposed method achieved a significant performance improvement on the \textit{CTraffic} dataset. We also conducted an experiment testing the performance of ReBoostSQL with LLMs, which are less capable than GPT-4 or ChatGPT. The results indicate that even with a less capable LLM, our method could still bring significant improvement in Text-to-SQL tasks.}

\section{Ackowledgement}

This work was conducted collaboratively among the authors. 

The authors would like to thank Dr. Xingyu Zeng, Dr. Feng Zhu, Kun Wang, Yuhang Ran, Zhiwei Xu, Tianpeng Bao, Shiwei Shi, and Guoqing Du for their valuable feedback, discussion, and participation in this project.

The authors would like to thank the product team for their efforts in preparing the query questions and high-quality database.

\bibliographystyle{IEEEtran}
\bibliography{ref}

\end{document}